\def\year{2019}\relax
\documentclass[letterpaper]{article} %DO NOT CHANGE THIS
\usepackage{aaai19}  %Required
\usepackage{times}  %Required
\usepackage{helvet}  %Required
\usepackage{courier}  %Required
\usepackage{url}  %Required
\usepackage{graphicx}  %Required
\usepackage{latexsym}
\usepackage{tabularx, booktabs}
\newcolumntype{Y}{>{\centering\arraybackslash}X}
\usepackage{xcolor}
\usepackage{soul}
\usepackage{amsmath}
\usepackage{multirow}
\usepackage{array}
\usepackage{subfigure}
\newcommand{\tabincell}[2]{
\begin{tabular}{@{}#1@{}}#2\end{tabular}
}

\def \B {\mathcal{B}}
\def \b {\mathbf{b}}
\def \c {\mathbf{c}}
\def \f {\mathbf{f}}
\def \h {\mathbf{h}}
\def \o {\mathbf{o}}
\def \p {\mathbf{p}}
\def \r {\mathbf{r}}
\def \v {\mathbf{v}}
\def \w {\mathbf{w}}
\def \x {\mathbf{x}}

\begin{document}
\title{Cross-relation Cross-bag Attention for  
Distantly-supervised Relation Extraction}

\author{
Yujin Yuan\textsuperscript{1},
Liyuan Liu\textsuperscript{2},
Siliang Tang\textsuperscript{1}\protect\thanks{Corresponding author.},
Zhongfei Zhang\textsuperscript{1},
Yueting Zhuang\textsuperscript{1},\\
{\bf \Large Shiliang Pu\textsuperscript{3},
Fei Wu\textsuperscript{1},
Xiang Ren\textsuperscript{4}}\\
\textsuperscript{1}{Zhejiang University}
\textsuperscript{2}{University of Illinois at Urbana Champaign}
\textsuperscript{3}{Hikvision Research Institute}\\
\textsuperscript{4}{University of Southern California}\\
\{yujin, siliang, zhongfei, yzhuang, wufei\}@zju.edu.cn,
llychinalz@gmail.com,\\
pushiliang@hikvision.com, 
xiangren@usc.edu}
\maketitle

\begin{abstract}
Distant supervision leverages knowledge bases to automatically label instances, thus allowing us to train relation extractor without human annotations. 
However, the generated training data typically contain massive noise, and may result in poor performances with the vanilla supervised learning.
In this paper, we propose to conduct multi-instance learning with a novel Cross-relation Cross-bag Selective Attention (C$^2$SA), which leads to noise-robust training for distant supervised relation extractor.
Specifically, we employ the sentence-level selective attention to 
reduce the effect of noisy or mismatched sentences,
while the correlation among relations were captured to improve the quality of attention weights.
Moreover, instead of treating all entity-pairs equally, we try to pay more attention to entity-pairs with a higher quality.
Similarly, we adopt the selective attention mechanism to achieve this goal.
Experiments with two types of relation extractor demonstrate the superiority of the proposed approach over the state-of-the-art, while further ablation studies verify our intuitions and demonstrate the effectiveness of our proposed two techniques.
\end{abstract}

\section{Introduction}
Aiming to detect and classify the relation between an entity pair in the given sentences, Relation Extraction (RE) plays a vital role in natural language understanding~\cite{etzioni2004web,mintz2009distant,liu2017heterogeneous}. 
The typical methods follow the supervised learning paradigm and require extensive human annotations, which are costly and time-consuming. 
To alleviate such reliance, attempts have been made to build relation extractor with distant supervision, i.e., automatically generating training data by knowledge base (KB).
For example, with the KB fact 
$<$\textit{Jimi Hendrix}, \texttt{died\_in}, \textit{London}$>$
in Table~\ref{tab:DS-RE exp}, distant supervision would annotate all sentences containing \textit{Jimi Hendrix} and \textit{London} as the relation type \texttt{died\_in}.

Despite its efficiency in cost and time, distant supervision is context-agnostic thus containing massive noise for sentence-level RE.
Accordingly, it may lead to an unsatisfying performance before noise-robust training paradigms are developed~\cite{lin2016neural}.
Multi-instance learning (MIL) has been employed to reduce the noise and bring such robustness. 
As shown in Table~\ref{tab:DS-RE exp}, it treats sentence bag as the basic training instance, instead of individual sentence.
Each bag would contain a group of sentences labeled by the same KB fact.  
By selecting from such bags, it allows a model to focus more on sentences of a higher quality and reduces the effect of noisy ones.
Specifically, some methods~\cite{riedel2010modeling,hoffmann2011knowledge,surdeanu2012multi} try to pick only one sentence from one bag,
while more improvements~\cite{lin2016neural,liu2017soft} have been observed by replacing the hard selection with Sentence-level Selective attention (ATT). 
ATT tries to assign attention weights to sentences and combines all sentences in the bag for the training.
  
\begin{table}[t]
    \centering
    \begin{footnotesize}
    \begin{tabular}{p{1.9cm}||p{0.4cm} p{4.6cm}}
    \hline
      KB Fact & \multicolumn{2}{c}{$<$\textit{Jimi Hendrix}, \texttt{died\_in}, \textit{London}$>$} \\ \hline \hline
    \multirow{3}{*}{\tabincell{l}{Sentence Bag \\ with Distant \\ Supervision}} & 
    S1 & \tabincell{l}{\textbf{Jimi Hendrix} died in 1970 in\\  \textbf{London} at 27.} \\
    \cline{2-3}
    & S2 & \tabincell{l}{George Frideric Handel and \textbf{Jimi} \\ \textbf{Hendrix} lived at adjacent addresses \\ in \textbf{London}} \\
    \cline{2-3}
    & ... & ...\\
    \hline
    \end{tabular}
    \caption{Distant Supervision and Sentence Bag.}
     \label{tab:DS-RE exp}
    \end{footnotesize}
\end{table}

However, ATT generates the attention weight for each relation type independently and overlooked their correlation. 
For example, in Table~\ref{tab:DS-RE exp}, by identifying S2 as a high-quality sentence for \texttt{live\_in}, we are also able to recognize it as a low-quality sentence for \texttt{die\_in}.
Based on this intuition, we propose a novel attention mechanism, Cross-relation Attention, which generates the attention weight after examining their relatednesses to all relation types.
 
Moreover, we go beyond ATT and 
construct training instances at a higher level. 
We relax the constraint that 
one training instance only contains one entity pair.
Specifically, we propose the Cross-bag Attention to combine different sentence bags, refer the
combined structure as superbag, and set superbag as the training instance instead of sentence bag. 
This allows us to focus more on sentence bags of a higher quality, and reduce the noise brought by KB Facts which are outdated or unexpressed in the corpus. 

Combining these two mechanisms, we refer our method as Cross-relation Cross-bag Selective Attention (C$^2$SA). 
Applying such attention to two types of relation extractor, we observe consistent improvements over the vanilla ATT.
Extensive ablation studies are further conducted to verify both our intuitions and the effectiveness of both components.

\section{Related work}
Relation extraction is one of the most important tasks in NLP.
Over the years, many efforts have been invested in relation extraction, 
especially in supervised relation extraction~\cite{mooney2006subsequence,zelenko2003kernel,rink2010utd}.
However, 
most of them are based on extra NLP systems to derive lexical features.
  
Recently, deep neural networks can learn underlying features automatically and have been used
in the literature.
\cite{socher2012semantic} uses a recursive neural network in relation extraction.
\cite{zeng2014relation,santos2015classifying,zeng2015distant} adopts an end-to-end 
convolutional neural network for relation extraction.
\cite{zhou2016attention,li2017multi,zhang2017position} uses the attention-based LSTM network
to mitigate the weakness of the CNN network in processing long-span information.
Based on CNN or RNN, there are still many efforts
\cite{xu2015semantic,xu2015classifying,vu2016combining,wang2016relation,jiang2016relation,huang2017deep}
to improve the 
network structures for more suitable for RE tasks. 
\cite{zeng2018large} trains a relation extractor using Reinforcement learning.
  
Although reasonably good performances are reported in the above models, 
training these models requires a large amount of annotation data, 
which are difficult and expensive to obtain.
To address this issue, distant supervision (DS) was proposed~\cite{mintz2009distant}
by assuming that all the sentences that mention two
entities of a fact triple describe the relation in the triple.
In order to suppress the large amount of noise introduced by DS,
many studies formulate the problem of relation classification as a multiple instance learning (MIL) problem
\cite{riedel2010modeling,hoffmann2011knowledge,surdeanu2012multi,zeng2015distant}.
All sentences containing the same entity pair are taken as a bag in MIL.
\cite{lin2016neural} proposes the selective attention to select high quality sentence features
in the bag as the bag feature and train the model by the bag feature.
\cite{luo3learning} proposes a transition matrix based method to dynamically characterize the noise.
\cite{feng2018reinforcement} uses reinforcement learning to select 
a more reliable subset on the DS dataset and uses it to train the classifier.
In order to solve the bag level noisy label problem, 
\cite{liu2017soft} uses a posterior probability constraint to correct potentially incorrect bag labels.
  
The selective attention method proposed by \cite{lin2016neural} is widely used in many recent efforts
\cite{liu2017soft,li2017multi,ji2017distant}.
The main differences between our approach and that selective attention are: 
1. Our approach takes into account the interplay between multiple relations.
2. Our approach assesses the quality of the bag feature and reduces the impact of bag-level noisy label problem while the existing
selective attention in the literature fails when processing a completely incorrect bag.

\section{Methodology}

Here, we develop a novel selective attention to reduce the noise of distant supervision for training relation extractors. 
We present the Cross-relation Cross-bag Selective Attention(C$^2$SA).
It improves the sentence-level attention by considering the correlation among relations, and conducts the selection at the bag level with another attention layer.

\begin{figure}[ht]
\includegraphics[width=1.0\linewidth]{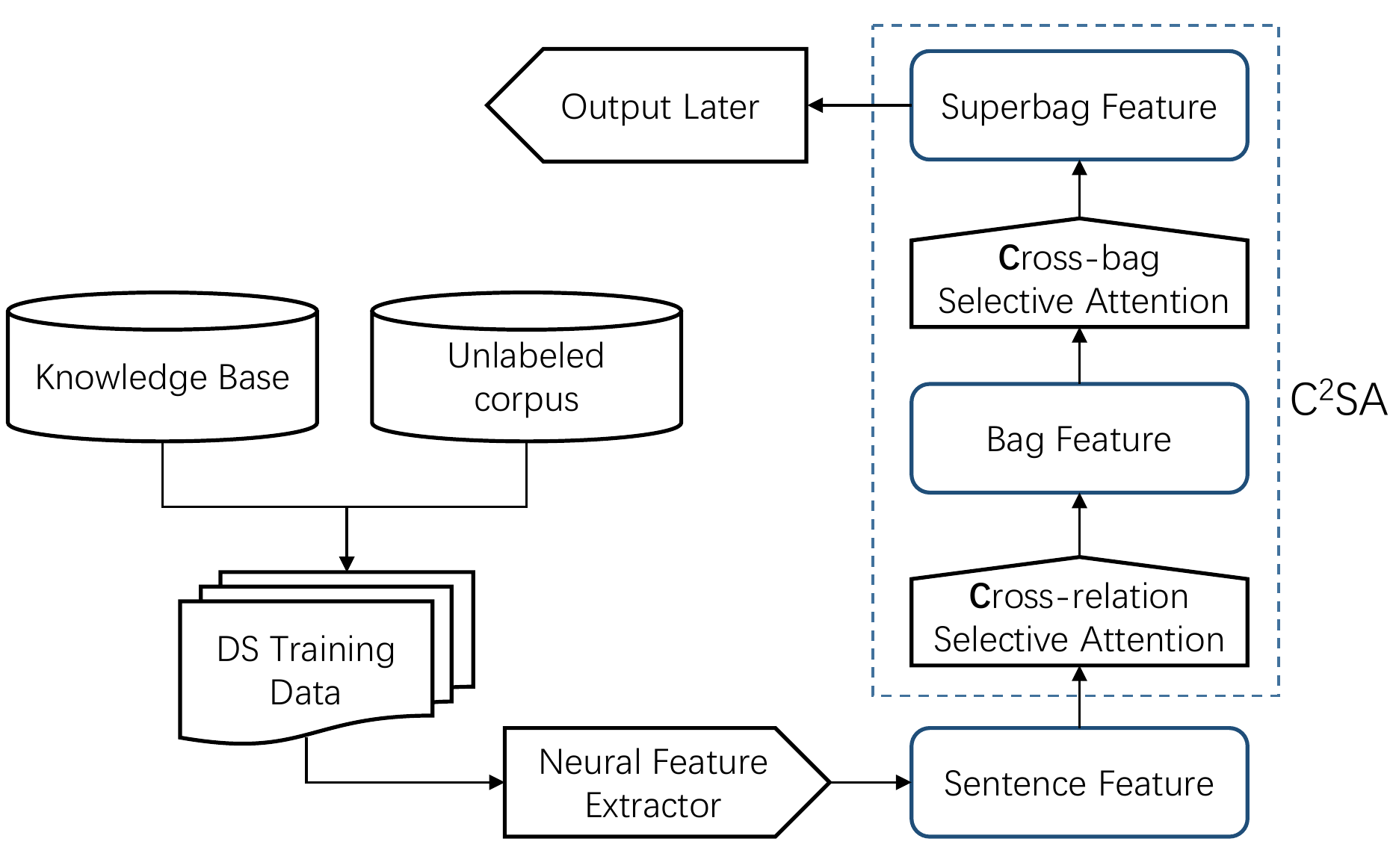}
\caption{Distant Training with C$^2$SA}
\label{overview}
\end{figure}

As in Figure~\ref{overview}, the relation extractor has two components
i.e., a neural feature extractor and an output layer.
As for the neural feature extractor, it extracts useful features for relation classification and can be embodied by any popular neural structures including CNNs and RNNs.
Based on the extracted features, the output layer makes the prediction on the relation type.

At the same time, the distant training pipeline with C$^2$SA has four steps. 
The neural feature extractor is first employed to construct representations for each sentence.
Then, the cross-relation selective attention combines the sentence representations and generate the representation for a sentence bag.
Similarly, the cross-bag selective attention combines representations for the sentence bags and generates the representation for the superbag.
At the end, the loss is calculated based on the superbag feature that guides the learning of the relation extractor.

We now proceed by introducing these components in further details.

\subsection{Relation Extractor}
Typically, the neural feature extractor can be considered as a neural sentence encoder, which encodes sentences into low-dimensional, fixed-length vectors.  
It can be employed as any neural encoder, such as RNNs (e.g., LSTMs, GRUs) or CNNs. 
Since the CNNs-based models achieve the best results in our experiments, we take them as the defaults.

\begin{figure}[t]
    \includegraphics[width=1.0\linewidth]{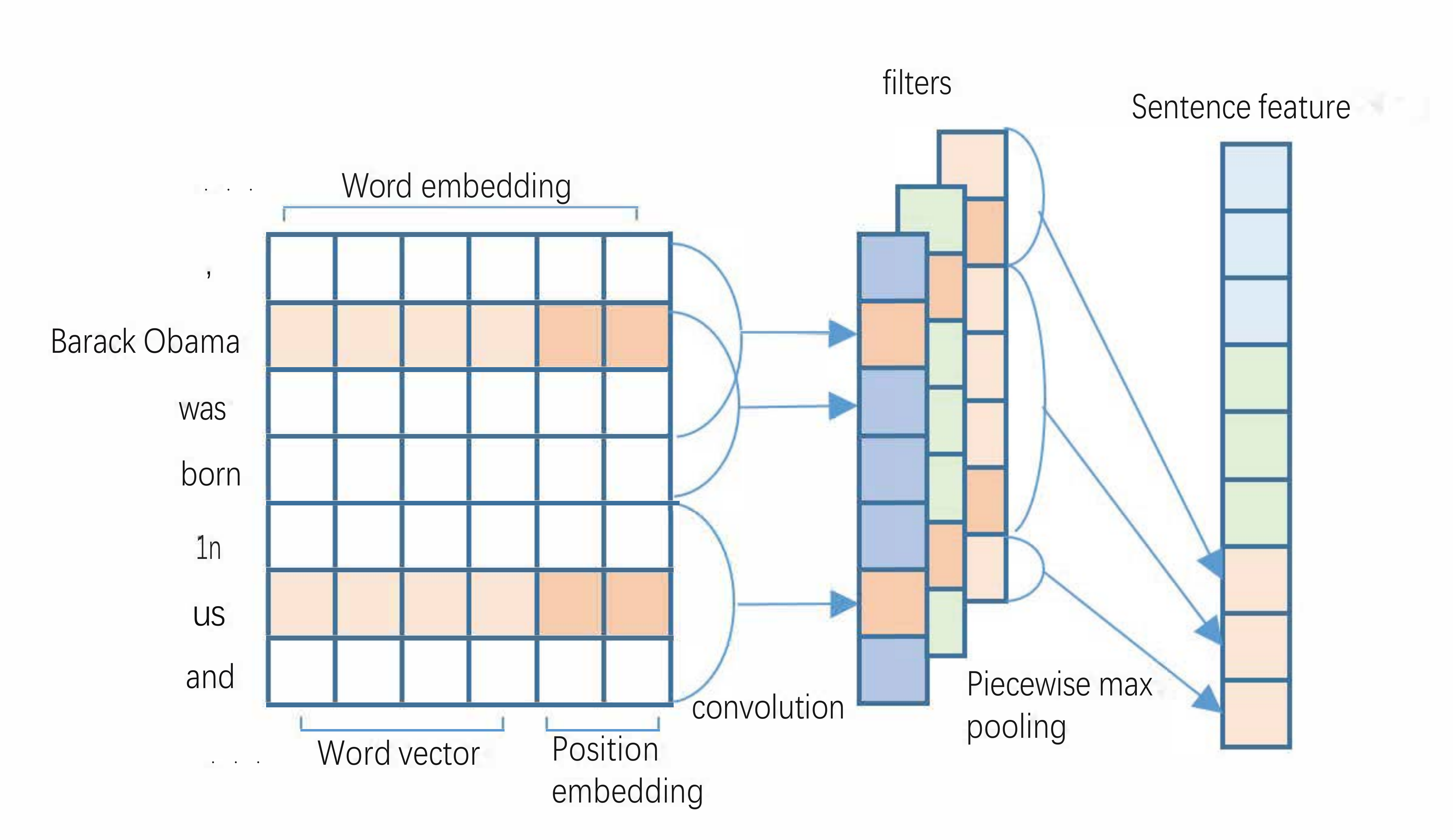}
    \caption{P-CNN based Neural Relation Extractor}
    \label{PCNN_structure}
\end{figure}
  
\subsubsection{Input Representation}
To encode the sufficient sentence information in an entity-aware manner, we formulate the input for neural networks as follows:
for a word at the position $i$ in a sentence,
we first transform it into a pre-trained word vector $\v_i$~ \cite{mikolov2013efficient}.
Then we calculate its relative distances to the target entities in the sentence (i.e., $d_1$ and $d_2$),
and find their position embedding vectors $\p_{d_1}$ and $\p_{d_2}$ by looking up a position embedding table~\cite{zeng2014relation}.
The position embedding table is randomly initialized and is updated during the model training.
After that, we concatenate $\v_i$ with $\p_{d_1}$ and $\p_{d_2}$ as the word representation $\w_i$.
By repeating these steps, we transform each sentence into a fixed-sized matrix $C=[\w_1,\w_2,\cdots,\w_m]^T$,
where $\w_i$ is a fixed-length vector and $m$ is the maximum length of a sentence in the whole data corpus. For shorter sentences, we pad them with zeros.
  
\subsubsection{Neural Feature Extractor}
As in Figure \ref{PCNN_structure}, we adopt piecewise-CNN (P-CNN) as the neural feature extractor.
It is composed of a Convolution layer and a Piecewise Max-pooling layer.
  
In the Convolution layer, the output $\c$ is calculated as:
\begin{equation*}
  \c_{i,j}=P_i \circ C_{j,j+l-1}.
\end{equation*}
where $P_i$ is the $i$-th convolutional kernel (filter), $l$ is the width of the kernel, and $C_{i,j}$ is a sliding window on sentence $C$ that starts from $\w_i$ and ends at $\w_j$, i.e., $C_{i,j} = \{\w_i, \w_{i+1}, \cdots, \w_j\}$.
  
The Piecewise Max-pooling~\cite{zeng2015distant} is a variant of the traditional max-pooling layer by considering the specific situation in relation extraction. 
For a sentence that contains an entity pair, the corresponding $\c$ is divided by such pair into three pieces. 
After that, the max-pooling operation is applied to each piece respectively, yielding three different output features.
We then concatenate them into one feature vector $\x_i$, where $\x_i \in R^{3\cdot n}$ and $n$ is the number of filters.
Finally, we apply hyperbolic tangent function at the output vector $\x_i$.
  
\subsubsection{Output Layer}
To compute the confidence of each relation, we employ the linear projection and softmax function to calculate the conditional probability:
  \begin{align}
    \o &= W\cdot \f \label{eq::linear}\\
    P(r|\f) &= \frac{e^{\o_r}}{\sum_{k=1}^{n_r}e^{\o_k}} \notag
  \end{align}
where $\f$ is the extracted feature and $W$ are the weights of the transformation. 
  
In the experiments, we adopt the dropout strategy~\cite{hinton2012improving} on the output layer to prevent overfitting.
Dropout prevents co-adaptation of hidden units by randomly setting them to zero for a proportion $p$. 
Thus, we revise Equation~\ref{eq::linear} to Equation~\ref{eq::linear_dropout}:
\begin{equation}
    \o=W \cdot (\f \odot \h) \label{eq::linear_dropout}
\end{equation}
where $\h$ is a vector of Bernoulli random variables with probability $p$ of being $1$.

\subsection{Cross-relation Cross-bag Selective Attention}
Now we describe the proposed Cross-relation Cross-bag Selective Attention for relation extractor training. 
As introduced before, we follow the standard MIL and construct the sentence bags, i.e.,  $B_i=\{x_{i, 1},x_{i, 2}, \cdots,x_{i, n_b}\}$, 
where $x_{i, *}$ contains the same entity pair and $n_b$ is the number of sentences in the bag.
With the distant supervision, each sentence bag is annotated by relations existing between the entity pair.
  
\begin{figure}[t]
    \includegraphics[width=1.0\linewidth]{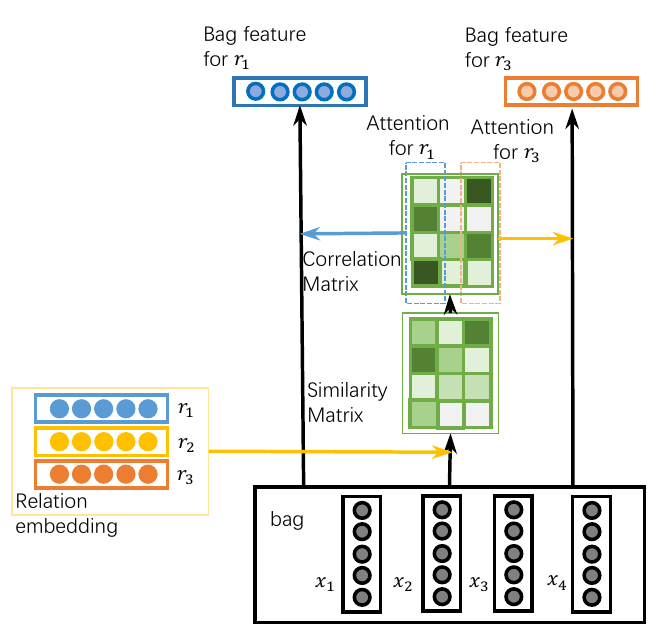}
    \caption{Structure of Sentence-level Cross-relation Selective Attention}
    \label{structure_of_SRA}
\end{figure}

With the sentence bag, we first leverage the cross-relation attention to combine sentences in the same bag (as shown in Figure~\ref{structure_of_SRA}); we then employ the cross-bag attention to integrate different bags into the superbag (as shown in Figure~\ref{Bag_level_attention}). 
The first attention attempts to reduce the effect of noisy or mismatched sentences, and the second aims to focus more attention on the high quality sentence bags.

\subsubsection{Cross-relation Selective Attention}
  
For sentence bag $B_i$, we refer the extracted feature representation for $x_{i, j}$ as to $\x_{i, j}$. Then we calculate the selective attention based on the similarity between the sentence and the relation:
\begin{equation}
    S_{i, j, k}=\frac{\x_{i, j} \cdot \r_k}{\left \| \x_{i, j} \right \|\left \| \r_k \right \|}
    \label{cosine_similarity}
\end{equation}
where $\r_k$ is the attention parameter corresponding to the $k$-th relation.
  
In order to capture the correlation among relations, we calculate the expected attention weight 
by the Bayes' rule:
\begin{align}
    & P(\mbox{ j}_{th} \mbox{sentence} | \mbox{ k}_{th} \mbox{relation}) \notag\\
    = & \frac{ P(\mbox{ k}_{th} \mbox{relation} | \mbox{ j}_{th} \mbox{sentence} ) P(\mbox{ j}_{th} \mbox{sentence})} {\sum_{\tilde{ j} = 1}^{n_b} P(\mbox{ k}_{th} \mbox{relation} | \tilde{\mbox{ j}}_{th} \mbox{sentence}) P(\tilde{\mbox{ j}}_{th} \mbox{sentence})}
    \label{eqn:bayes}
\end{align}
  
Specifically, we assume
  $P(\mbox{ j}_{th} \mbox{sentence})$ to be the uniform distribution, and calculate $P(\mbox{ k}_{th} \mbox{relation} | \mbox{ j}_{th} \mbox{sentence} )$ with the softmax function:
\begin{equation}
    P(\mbox{ k}_{th} \mbox{relation} | \mbox{ j}_{th} \mbox{sentence} ) = \frac{e^{S_{i, j, k}}}{\sum_{\tilde{ k}=1}^{n_r}e^{S_{i, j, \tilde{ k}}}}
    \label{gen_alpha}
\end{equation}
  
To simplify the notion, we refer the calculated value of $P(\mbox{ k}_{th} \mbox{relation} | \mbox{ j}_{th} \mbox{sentence} )$ as to $\alpha_{j, k}$, and the value of $P(\mbox{ j}_{th} \mbox{sentence} | \mbox{ k}_{th} \mbox{relation})$ as to $\beta_{j, k}$. 
Then we rewrite Equation~\ref{eqn:bayes} as 
\begin{equation}
    \beta_{j, k}=\frac{{\alpha_{j, k}}}{\sum_{\tilde{ j} = 1}^{n_b}{\alpha_{\tilde{ j},k}}}
    \label{gen_beta}
\end{equation}
  
Accordingly, the bag feature for $B_i$ for the k-th relation can be calculated as 
\begin{equation}
    \b_{i, k} = \sum_{\tilde{ j}=1}^{n_b} \beta_{\tilde{ j}, k} \x_{i, \tilde{ j}}
    \label{gen_bag_feature}
\end{equation}
  
\begin{figure}[t]
      \includegraphics[width=1.0\linewidth]{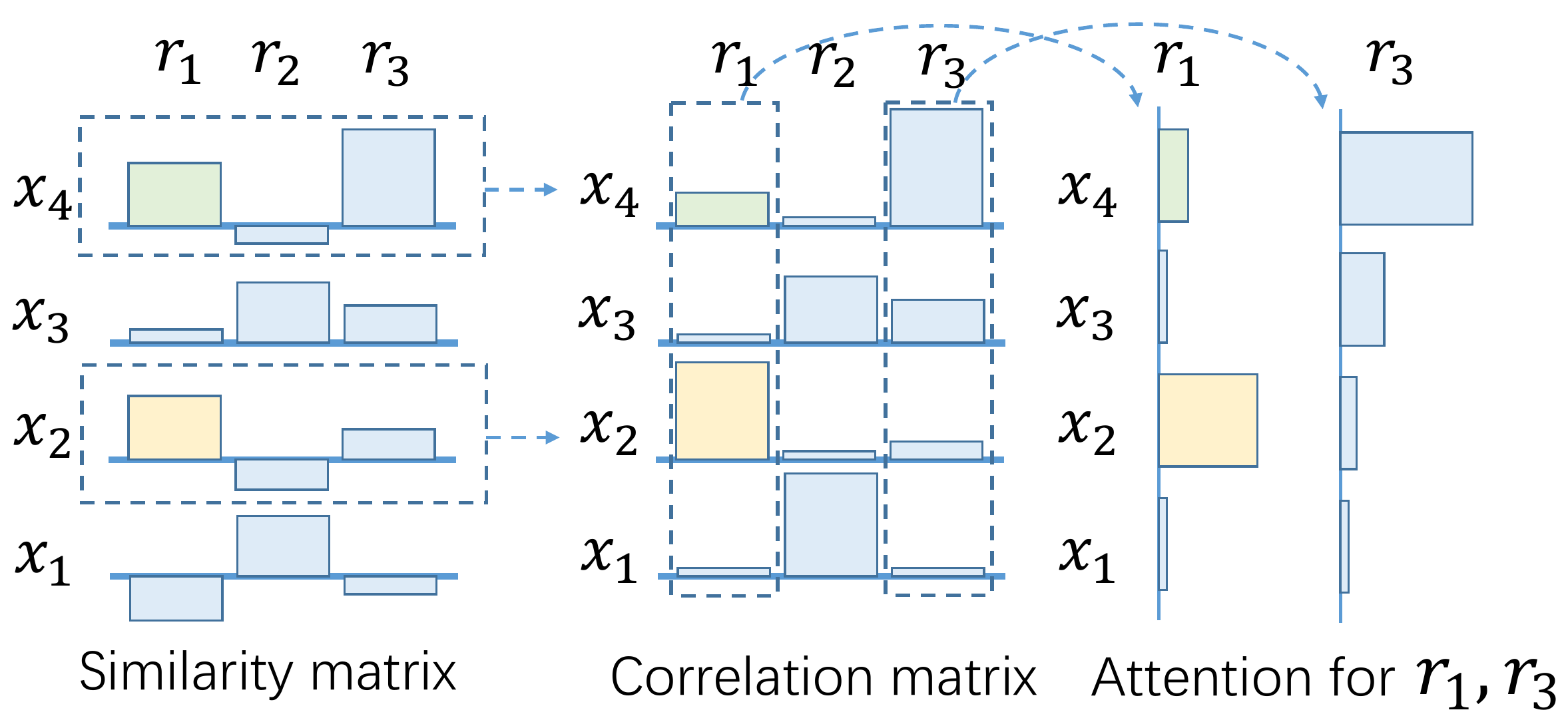}
      \caption{Similarity matrix and Correlation matrix}
      \label{matrix}
\end{figure}
  
 As shown in Figure~\ref{matrix}, the cross-relation selective attention depends on the similarities from the sentence to not only the target relation, but also the other relations. 
For example, in Figure~\ref{matrix}, 
$x_2$ and $x_4$ have similar similarities to $r_1$, 
but since $x_4$ is more inclined to expressing $r_3$, 
the model tends to use features of $x_2$ to generate the bag feature and use this feature to predict $r_1$.
 
\subsubsection{Cross-bag Selective Attention}
The sentence level attention assumes that at least one sentence in a bag expresses the relation between entity pairs.
However, due to the noisy nature of distant supervision, there may still exist noise in the sentence bag level.
For example, there are a large number of entity pairs that cannot find any expression about their relations in the given corpus.
Such entity pairs would result in the mismatched and noisy training instances for the sentence level relation extraction.
  
\begin{figure}[t]
    \includegraphics[width=1.0\linewidth]{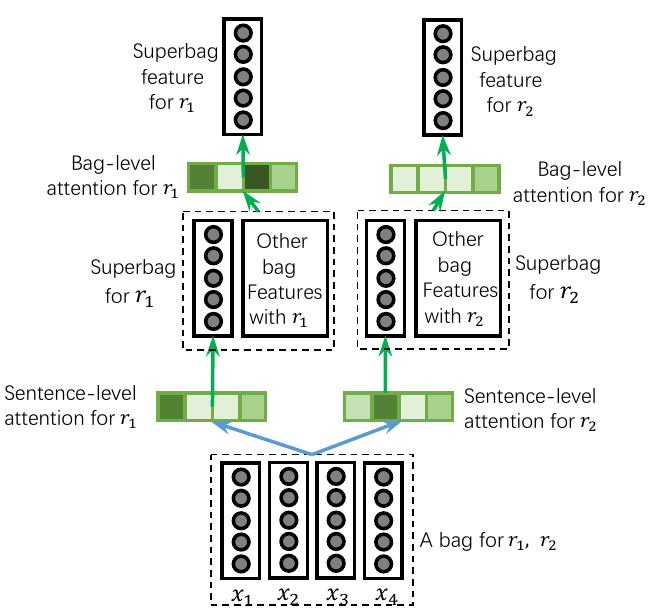}
    \caption{Cross-bag Selective Attention}
    \label{Bag_level_attention}
\end{figure}

As shown in Figure~\ref{Bag_level_attention}, we go beyond the existing setting and propose to establish an additional selective attention layer, i.e., Cross-bag Selective Attention.
Specifically, we intend to combine several sentence bags of the same relation type and 
put more attention to the higher quality ones.
We refer the higher-level structure that contains a group of sentence bags as to {\em superbag}, and denote it as $\B=\{B_{1},B_{2},...,B_{n_s}\}$, where $n_s$ is the size of the superbag and all $B_{i}$ are labelled with the $k$-th relation type.
  
Based on the Cross-relation Selective Attention, we construct the representation for each sentence bag while capturing the correlation among sentences.
Here, we combine these representations with an attention layer.
Specifically, we obtain the superbag feature $\f$ for $\B$ as:
\begin{align}
    \f&=\sum_{i=1}^{n_s} \gamma_i \cdot \b_{i, k} \notag \\
    \gamma_i &= \frac{e^{S(\r_k, \b_{i, k})}}{\sum_{j=1}^{n_s}e^{S(\r_k, \b_{j, k})}}
    \label{SB}
\end{align}
where $\b_{i, k}$ is the bag representation w.r.t. $B_i$ for the k-th relation and $\r_k$ is the attention parameter corresponding to the $j$-th relation.
Specifically, we tie up the $\r_k$ in Equation~\ref{SB} with the ones in Equation~\ref{cosine_similarity}.
Also, similar to the Equation~\ref{cosine_similarity}, we calculate $S(\r_k, \b_{i, k})$ with the cosine similarity.
  
\subsection{Model Training}
We employ the objective function as the negative log likelihood at the superbag level as follows:
\begin{equation}
    \mathcal{L}=-\sum_{i=1}^{n_{sb}}log p(l_i|\B_i)
\end{equation}
where $n_{sb}$ indicates the number of the superbags in the training set and
$l_i$ indicates the label of a superbag.

\section{Experiments}
  
\begin{table}[t]
\center
\begin{tabular}{c|c|c}
\hline
Parameter Name & Value & Candidate Set \\ \hline \hline
sentence embedding & \multirow{2}{*}{100} & \multirow{2}{*}{\{100, 150, 200\}}\\
dimension & & \\\hline
batch size & 100 & \{100, 150, 200\} \\\hline
superbag size & 3 & \{2, 3, 4, 5\} \\\hline \hline
sliding window size & 3 & \multirow{5}{*}{\tabincell{c}{reused \\ from \\ previous \\ work}}\\\cline{1-2}
word vector dimension & 50 & \\\cline{1-2}
position embedding & \multirow{2}{*}{5} & \\
dimension & & \\ \cline{1-2}
dropout probability & 0.5 & \\ \hline
\end{tabular}
\caption{Hyper-parameter Settings.}\label{tbl:hyper}
\end{table}

We report the performances of C$^2$SA by comparing it with the state-of-the-art relation extraction methods.
Some of these baselines are BLSTMs-based, while others are P-CNNs-based.
Since C$^2$SA is model-agnostic and only used in the learning phrase, we conduct the experiments with both types of neural relation extractors.
Moreover, we employ two evaluation settings for an extensive comparison. 
Specifically, we first follow the popular setting and evaluate the model performances on the corpus-level relation extraction task.
Besides, we also conduct experiments on the sentence-level relation extraction task with a human annotated test corpus.
We observe that, in all settings, our proposed C$^2$SA consistently outperforms the state-of-the-art.

For a better understanding, we further report a case study and ablation experiments, which further verifies our intuition and demonstrates the effectiveness of both cross-sentence and cross-bag selective attention mechanisms.

\subsection{Model Training}
As discussed before, we leverage C$^2$SA to train two types of relation extractors, i.e., P-CNNs-based and BLSTMs-based.
We refer these two variants as PCNN+C$^2$SA and BLSTM+C$^2$SA.

\subsubsection{Dataset}
Following the existing literature~\cite{riedel2010modeling,lin2016neural,li2017multi,feng2018reinforcement,liu2017soft}, we use the New York Times (NYT) dataset as the training set~\cite{mintz2009distant}.
It uses Freebase~\cite{bollacker2008freebase} to provide distant supervision on the NYT corpus.
Specifically, it collects sentences from 2005 to 2006 and supports 53 different relations (including NA which means no relations for an entity pair). 
For training set statistics, this dataset contains 522611 sentences, 281270 entity pairs and 18252 KB facts.
  
\subsubsection{Model Setting}
In all the experiments, we use 50 dimensional word vectors that are pre-trained by the Skip-gram algorithm \footnote{https://code.google.com/p/word2vec/} on the NYT corpus.
For hyper-parameters, we reuse part of them from the previous study~\cite{zeng2015distant,li2017multi}, and tune the rest part by grid-search with the three-fold cross-validation (on the training set).
The final hyper-parameter setting used in our experiments are summarized in Table~\ref{tbl:hyper}.
  
\begin{figure}[t]
    \includegraphics[width=\columnwidth]{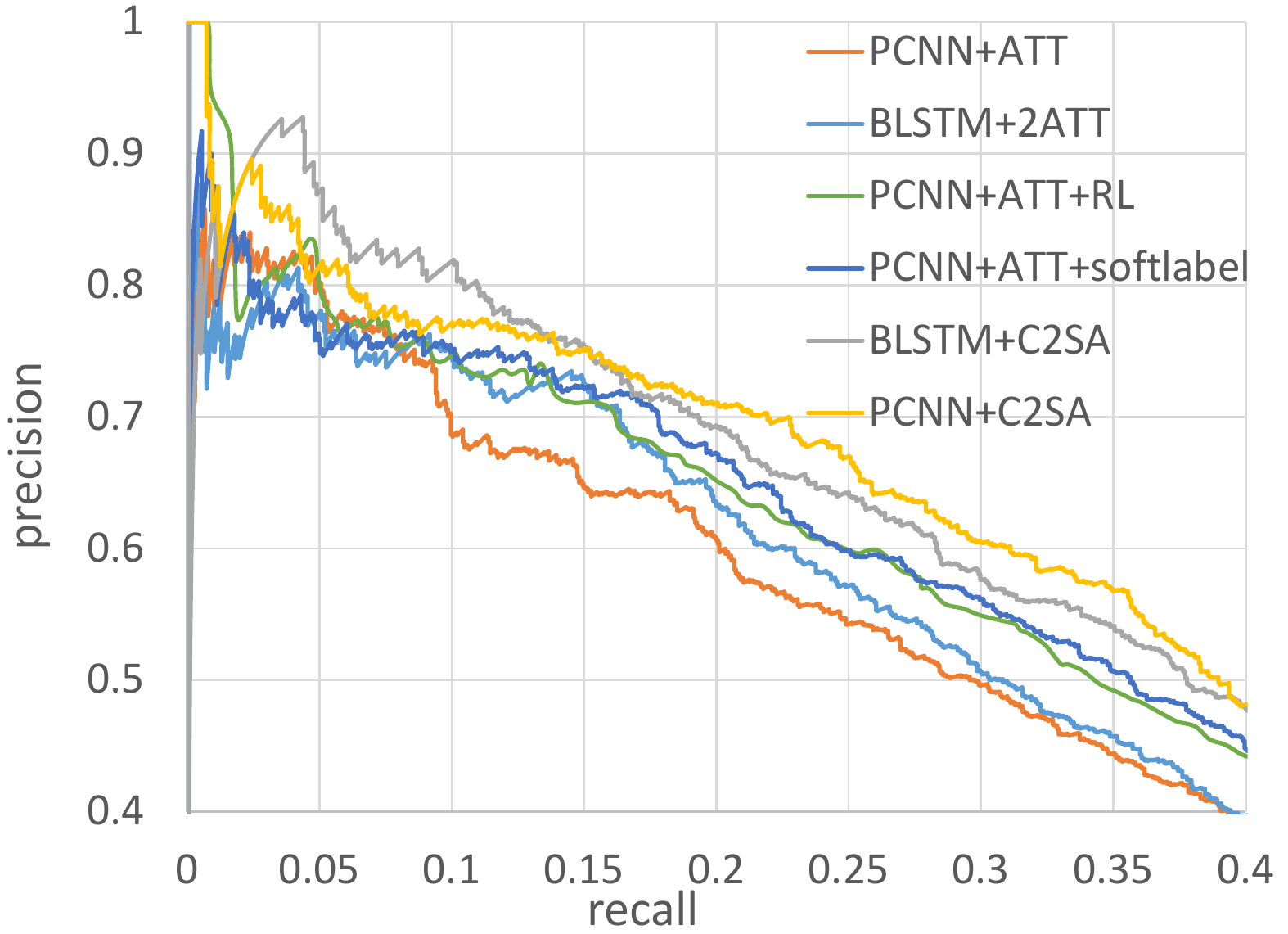}
    \caption{Performance comparison on the corpus-level relation extraction.}
    \label{Main_result}
\end{figure}
  
\subsection{Comparison on the Corpus-level Task}
Here, we evaluate the performances of our method on the corpus-level relation extraction. 
For an entity pair, the task is to identify their relation type with regard to all the sentences that mention this entity pair.
Specifically, we first feed all the sentence representations to the output layers for prediction. 
After that, the probability of a relation for an entity pair is the maximum probability of this relation in all the sentences mentioning this entity pair.
  
\subsubsection{Test Set}
We use the test set of NYT for the corpus-level evaluation~\cite{mintz2009distant}.
The test set is constructed with sentences from NYT of 2007 annotated with Freebase~\cite{bollacker2008freebase}.
Specifically, it contains 172448 sentences, 96678 entity pairs and 1950 KB facts.
  
\subsubsection{Competing Methods}
We choose four recent methods as baselines:
\begin{itemize}
\item PCNN+ATT~\cite{lin2016neural} uses the vanilla sentence-level selective attention to combine sentence features for each bag. Based on this attention, the representation for each bag is obtained and is trained under the PCNN model.
\item BLSTM+2ATT~\cite{li2017multi} also employs the vanilla sentence-level selective attention. Different from PCNN+ATT, it is a BLSTM-based model which has an additional word-level attention module.
\item PCNN+ATT+RL~\cite{feng2018reinforcement} further incorporates reinforcement learning to improve PCNN+ATT. It trains the PCNN+ATT model with a subset of the training set, which is selected by the learned policy.
\item PCNN+ATT+softlabel~\cite{liu2017soft} enhances the PCNN+ATT model by using the posterior probability constraint to correct potentially incorrect bag labels.
\end{itemize}
  
\subsubsection{Performance Comparison}
Following the existing literature~\cite{riedel2010modeling,lin2016neural,li2017multi,feng2018reinforcement,liu2017soft}, we evaluate C$^2$SA in the held-out setting and report the model performances with the Precision-Recall curve. 

\begin{figure*}[t]
    \centering
    \subfigure[]{
      \begin{minipage}{8.5cm}
      \centering
      \includegraphics[width=1.0\linewidth]{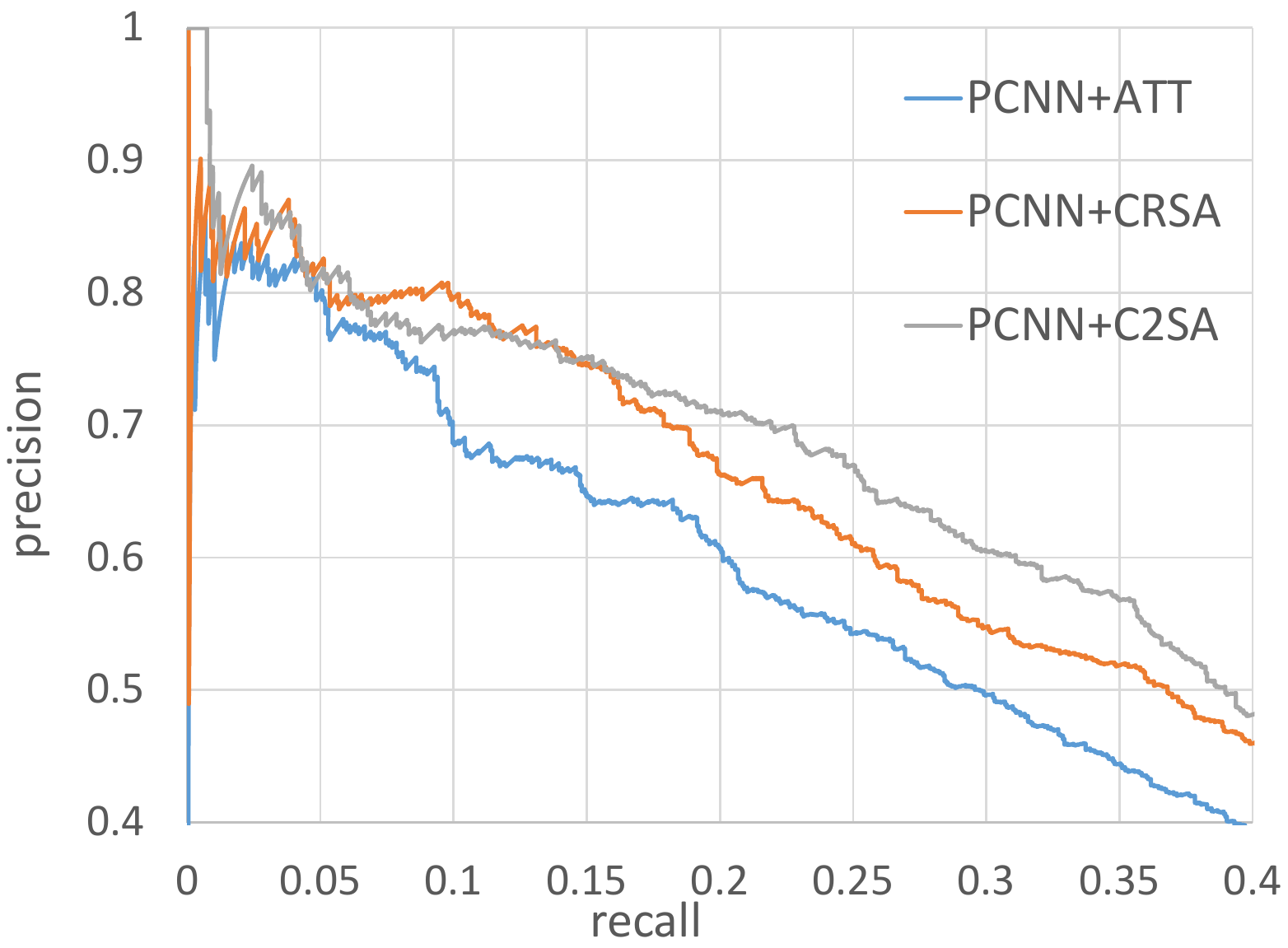}
      \end{minipage}
    }
    \subfigure[]{
      \begin{minipage}{8.5cm}
      \centering
      \includegraphics[width=1.0\linewidth]{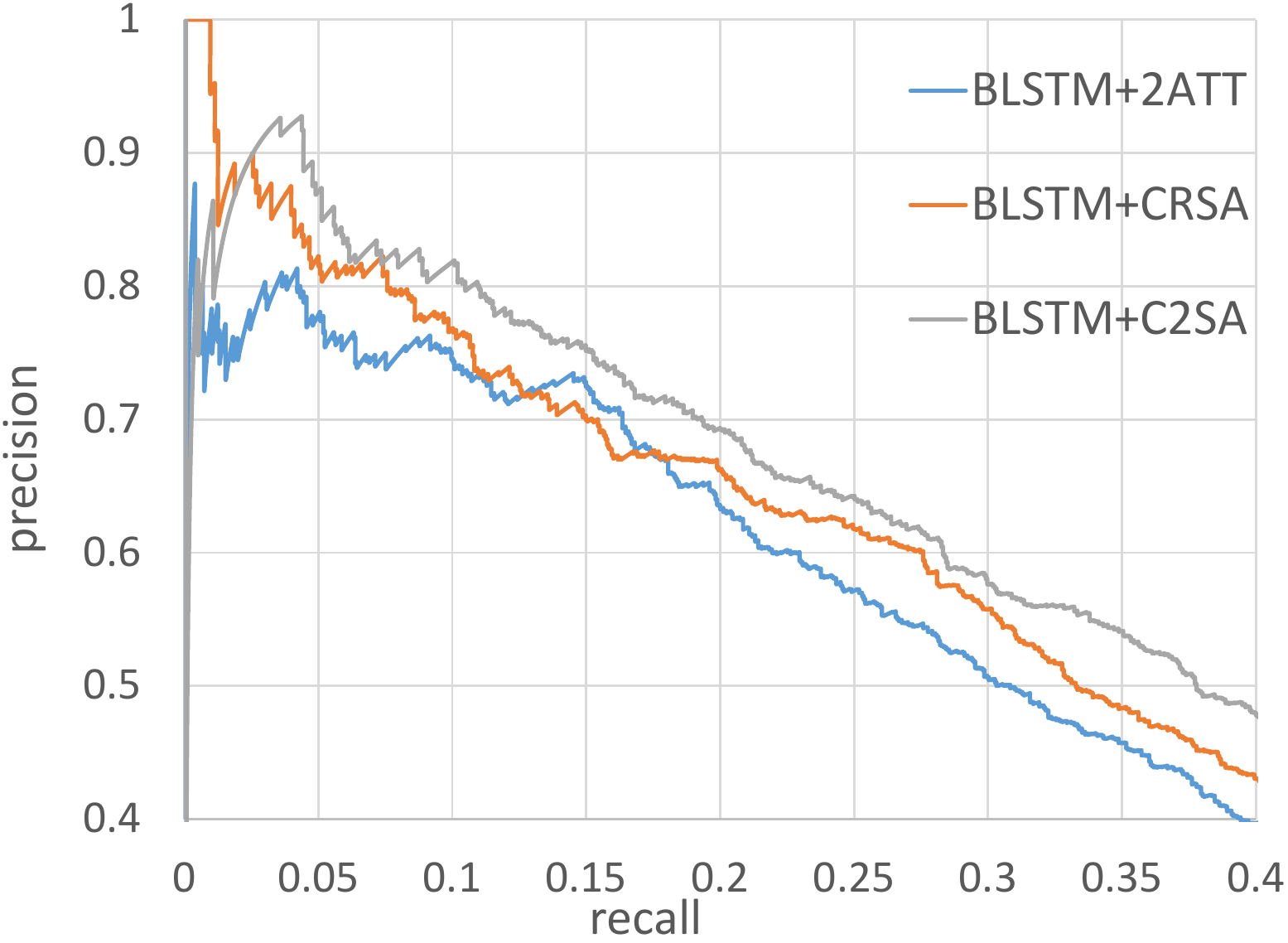}
      \end{minipage}
    }
    \caption{(a) Aggregate precision/recall curves of PCNN+ATT, PCNN+CRSA, PCNN+C$^2$SA
    (b) Aggregate precision/recall curves of BLSTM+2ATT, BLSTM+CRSA, BLSTM+C$^2$SA}
      \label{exp_results}
\end{figure*}
  
We summarize the performances of BLSTM+C$^2$SA, PCNN+C$^2$SA, and all the baselines in Figure~\ref{Main_result}.
We observe that models trained with C$^2$SA, no matter whether they are PCNNs-based or BLSTMs-based, consistently outperform the other models.
In addition, there is no clear difference between PCNN-based methods and BLSTM-based methods.
For example, the PCNNs-based baseline (PCNN+ATT) performs worse than the BLSTMs-based baseline (BLSTM+2ATT).
However, with our proposed C$^2$SA, the PCNNs-based method (PCNN+C$^2$SA) outperforms the BLSTMs-based method (BLSTM+C$^2$SA).
We report the ablation studies later.
  
At the same time, we observe that the choice of the training setting has a large impact on the performances.
That is, with the model trained more robustly to noise, there are clear improvements in performances.
These observations have further verified our intuition for improving the distantly-supervised relation extraction with a more robust training.
  
\subsection{Comparison on the Sentence-level Task}
For an extensive comparison, we further evaluate the performances of our method on the sentence-level relation extraction. Different from the corpus-level task, this task aims to identify the relation type for an entity pair with regard to a specific sentence.
More precisely, we feed the sentence representation to the output layer and observe the prediction.
  
\subsubsection{Test Set}
For the sentence-level evaluation, we adopt the dataset used in the existing literature ~\cite{hoffmann2011knowledge}.
It contains 395 sentences with human annotations.
Compared with the test set for the corpus-level task, this set is small in size.
However, considering the fact that this dataset is manually annotated, it makes sense to report this comparison study.
  
\subsubsection{Competing Methods}
We compare C$^2$SA with two major baselines, i.e., PCNN+ATT and BLSTM+2ATT.
Besides, we also make comparison with two variants of C$^2$SA, i.e., CRSA and C$^2$SA-dot.
CRSA only performs the cross-relation selective attention and trains the model at the sentence bag level.
C$^2$SA-dot, on the other hand, changes the scoring function in Equation~\ref{cosine_similarity} and Equation~\ref{Bag_level_attention} from the cosine similarity to the dot product.
  
\subsubsection{Performance Comparison}
As reported in Table~\ref{manual-annotation-test}, we observe that C$^2$SA outperforms both baselines and the variants.
Besides, we observe that CRSA outperforms ATT and C$^2$SA outperforms CRSA. 
This observation verifies the effectiveness of the two proposed techniques.
In addition, we observe that after replacing the cosine similarity with dot product in C$^2$SA, the performance drops significantly.
This shows that when serving as the scoring function for the selective attention, cosine similarity is more effective than the dot product.
  
\begin{table}[t]
\begin{center}
\begin{tabularx}{\columnwidth}{r *{1}{Y} r *{1}{Y}}
\toprule
PCNNs-based & F1 & BLSTMs-based & F1\\\midrule \midrule
PCNN+ATT &  0.377  & BLSTM+2ATT & 0.382\\\midrule
PCNN+CRSA &  0.411  & BLSTM+CRSA & 0.409\\\midrule
PCNN+C$^2$SA &  0.421  & BLSTM+C$^2$SA & 0.448\\\midrule
\tabincell{r}{PCNN\\+C$^2$SA-dot} &  0.400  & \tabincell{r}{BLSTM\\+C$^2$SA-dot} & 0.401\\
\bottomrule
\end{tabularx}
\end{center}
\caption{Performance comparison on the sentence-level relation extraction.}
\label{manual-annotation-test}
\end{table}

\subsection{Ablation Study}
To further demonstrate the effectiveness of the two proposed selective attention mechanisms, we now report the ablation studies.
  
To examine the effectiveness of the cross-relation selective attention, we compare CRSA with the vanilla selective attention (ATT).
Specifically, CRSA only performs the cross-relation selective attention; both CRSA and ATT conduct the training at the sentence bag level.
As to the cross-bag selective attention, it is based on the sentence bag representation.
In other words, only the sentence-level attention is required.
Accordingly, we demonstrate the effectiveness of the cross-bag selective attention by comparing CRSA with the full C$^2$SA. 
  
For a fair comparison, we compare the two types of relation extractions separately as we did before. 
As shown in Figure~\ref{exp_results}, all the models are trained with the same training set introduced before and are evaluated on the corpus-level task. 
We employ the Precision-Recall curve for the comparison studies with the summarized performances of PCNNs-based relation extractions in Figure~\ref{exp_results}(a) and the summarized performances of  BLSTMs-based ones in Figure~\ref{exp_results}(b).
  
We observe that C$^2$SA achieves the best performance in both cases.
Also, we note that the PCNNs-based model achieves a better improvement than the BLSTMs-based model.
For example, there is an obvious margin between the PCNN+CRSA and PCNN+ATT.
We believe that this observation is due to the difference between the characteristics of the different neural feature extractors.
The CNNs-type models are good at extracting local information (such as trigger words) of a sentence 
reflected in the dimensions of their feature vectors, leading to the observation that cosine similarity based attention mechanism delivers a better performance.

On the other hand, it is observed that the gap between C$^2$SA and CRSA verifies the fact that some sentence bags are of a higher quality than others.
This observation demonstrates that the proposed cross-bag selective attention allows the model training at the superbag level with a more robust performance.
We report the case studies below to support this observation.
  
\begin{table}[t]
\begin{center}
\begin{tabularx}{\columnwidth}{r *{1}{Y}}
\toprule
Total count of sentence bags & 100 \\\midrule
Contain at least one sentence & 69 \\\midrule
All annotations are incorrect & 31 \\
\bottomrule
\end{tabularx}
\end{center}
\caption{Manually checked qualities for the sampled sentence bags with distant supervision}
\label{Dataset}
\end{table}

\subsection{Case Study}
  
\subsubsection{Noise of Distant Supervision at Bag-level}

Now we report the case studies to verify the intuition about the cross-bag selective attention and superbag learning.
We randomly select 20 different relation types from the NYT dataset, randomly select 100 entity pairs, construct 100 sentence bags, and manually examine their qualities.
Specifically, these sentence bags contain 483 sentences and their qualities are summarized in Table~\ref{Dataset}.

We observe that, even after aggregating sentences to construct sentence bag, the distant supervision is still noisy. 
Actually, about 31\% of all the sentence bags do not even contain one sentence that is correctly annotated.
Therefore, it is fair to conclude that the superbag-level training is necessary to handle the noise of distant supervision.

\begin{table}[t]
\centering
\begin{footnotesize}
\begin{tabular}{p{1.2cm}||p{3cm}|p{1.15cm}|p{1.15cm}}
\hline
& \multicolumn{3}{c}{Superbag label: person-place\_lived} \\
\hline \hline
KB-Facts & \tabincell{l}{Sentences} & \tabincell{l}{Sentence\\attention} & \tabincell{c}{Bag\\attention} \\
\hline \hline
\multirow{7}{*}{\tabincell{l}{\textit{Hunter\_}\\\textit{s.\_thom}\\\textit{-pson},\\ \texttt{live}\\\texttt{d\_in},\\ \textit{Colo}\\\textit{-rado}}} & \tabincell{l}{\textbf{Hunter\_s.\_thompson} \\ who committed \\ suicide  last month \\ in \textbf{Colorado} ...\\} & low & \multirow{5}{*}{high} \\
\cline{2-3}
& \tabincell{l}{\textbf{Hunter\_s.\_thompson}\\ lived and wrote here in \\ the high rocky \\ mountains of \\ central \textbf{Colorado} ...\\} & high  & \\
\hline
\hline
\multirow{5}{*}{\tabincell{l}{\textit{Dan\_b}\\\textit{artlett},\\ \texttt{live}\\\texttt{d\_in},\\ \textit{Texas}}} & \tabincell{l}{President is residing \\ in \textbf{Texas}, ''said \\ \textbf{Dan\_bartlett}, \\ counselor to \\ the president.} & medium & \multirow{4}{*}{low} \\
\cline{2-3}
& \tabincell{l}{\textbf{Dan\_bartlett}, president \\ Bush`s counselor,\\ said ... in \textbf{Texas} ...} & medium  & \\
 \hline
\end{tabular}
\caption{A case study for two sentence bags and their corresponding superbag}
 \label{case_studies}
\end{footnotesize}
\end{table}

\subsubsection{Effectiveness of Cross-bag Selective Attention}
We further examine whether the proposed cross-bag selective attention is capable of properly handling the noise from distant supervision.
Table~\ref{case_studies} shows a superbag for the relation type \texttt{lived\_in}.
It contains two sentence bags. One is for the KB fact $<$\textit{Hunter\_s}, \texttt{lived\_in}, \textit{Colorado}$>$, and the other is for the KB fact $<$\textit{Dan\_bartlett}, \texttt{lived\_in}, \textit{Texas}$>$.
We observe that there is only one sentence that is correctly annotated in the first sentence bag, and that all the sentences in the second bag are unmatched with the relation type of the KB fact. 
Clearly, the sentence-level selective attention would fail to handle the second bag.
On the other hand, the proposed cross-bag selective attention allows the model to focus more on the sentence bag with a higher quality.
For example, in Table~\ref{case_studies}, such selective attention reduces the effect of the distant supervision noise by assigning a smaller attention weight to the second sentence bag.

\section{Conclusion and Future Work}
In this paper, we have proposed the Cross-relation Cross-bag Selective Attention to develop a better relation-entity model to effectively learn the features of true expression relations from typically noisy distant supervision data. Experiments show that the proposed attention model is capable of learning higher quality bag features than the existing literature. In addition, it is also demonstrated that Cross-bag Selective Attention is further capable of boosting the performances through the high-quality bag features. 

\section{Acknowledgments}
This work has been supported in part by NSFC (No. 61751209, U1611461), 973 program (No. 2015CB352302), Hikvision-Zhejiang University Joint Research Center, Chinese Knowledge Center of Engineering Science and Technology (CKCEST), Engineering Research Center of Digital Library, Ministry of Education. Xiang Ren's research has been supported in part by National Science Foundation SMA 18-29268.
 
\fontsize{9.5pt}{10.5pt}
\selectfont
\bibliography{C2SA_reference}
\bibliographystyle{aaai}
\end{document}